\newif\ifdraft\draftfalse

\newif\iffinal\finalfalse
\finaltrue

\newif\ifinlineref\inlinereffalse
\inlinereftrue

\newif\ifdotikz\dotikzfalse

\newif\ifreva\revafalse

\newif\ifrevb\revbfalse

\documentclass[oneside,a4paper]{article}

\usepackage[sort]{natbib}
\newcommand{\shortcite}[1]{(\citeyear{#1})}
\usepackage{fullpage}
\usepackage[utf8]{inputenc}
\usepackage[T1]{fontenc}
\usepackage{graphics}
\usepackage{paralist}
\usepackage{url}
\usepackage{fancyvrb}
\usepackage{booktabs}
\usepackage{amsmath}
\usepackage{amssymb}

\ifreva
\usepackage{color}
\newcommand{\revamark}[0]{\color{blue}}
\else
\newcommand{\revamark}[0]{}
\fi

\ifrevb
\usepackage{color}
\newcommand{\revbmark}[0]{\color{blue}}
\newcommand{\revb}[1]{{\revbmark{}#1}}
\else
\newcommand{\revbmark}[0]{}
\newcommand{\revb}[1]{{#1}}
\fi

\ifdotikz
\usepackage{tikz}
\usetikzlibrary{shapes,arrows,backgrounds,%
matrix,patterns,arrows,decorations.pathmorphing,decorations.pathreplacing,%
positioning,fit,calc,decorations.text,shadows%
}
\pgfrealjobname{cr}
\else
\long\def\beginpgfgraphicnamed#1#2\endpgfgraphicnamed{\includegraphics{#1}}
\fi

\usepackage{mymacros}

\title{Marmara Turkish Coreference Corpus and Coreference Resolution Baseline}

\author%
{Peter SCHÜLLER$^{1,2,\star}$, Kübra CINGILLI$^1$, Ferit TUNÇER$^1$, \\
Barış Gün SÜRMELİ$^{1,3}$, Ayşegül PEKEL$^1$, \\
Ayşe Hande KARATAY$^{1,4}$, Hacer Ezgi KARAKAŞ$^5$ \\
$^1$Marmara University \\
Faculty of Engineering \\
Istanbul, Turkey \\
$^2$Technische Universit{\"{a}}t Wien \\
Institut f{\"{u}}r \revb{Logic and Computation} \\
Knowledge-Based Systems Group \\
Vienna, Austria \\
$^3$ inIT - Institut für industrielle Informationstechnik \\
Hochschule Ostwestfalen-Lippe \\
Lemgo, Germany \\
$^4$Technische Universit{\"{a}}t M{\"{u}}nchen \\
Fakult{\"{a}}t f{\"{u}}r Informatik \\
Garching, Germany \\
$^5$Sabanci University \\
Faculty of Engineering and Natural Science \\
Istanbul, Turkey \\
$^\star$Corresponding Author \\
{\tt schueller.p@gmail.com} \quad {\tt ps@kr.tuwien.ac.at} \\
{\tt kubracingilli@gmail.com} \quad
{\tt ferit@cryptolab.net} \\
{\tt baris.suermeli@hs-owl.de} \quad
{\tt aysegulpekel@gmail.com} \\
{\tt handekaratay7@gmail.com} \quad
{\tt hezgikarakas@sabanciuniv.edu}
}

\date{~\\[1em]
Version 2, \today\\[1em]
Technical Report: manuscript submitted to\\
\emph{Natural Language Engineering} by Cambridge University Press \\
\url{http://journals.cambridge.org/nle}}

\begin{document}
\label{firstpage}
\maketitle

\begin{abstract}
We describe the Marmara Turkish Coreference Corpus, which is an annotation of the whole METU-Sabanci Turkish Treebank with mentions and coreference chains. Collecting eight or more independent annotations for each document allowed for fully automatic adjudication. We provide a baseline system for Turkish mention detection and coreference resolution and evaluate it on the corpus.
\end{abstract}

\section{Introduction}

Coreference Resolution is the task of identifying groups of
phrases in a text that refer to the same discourse entity.
Such referring phrases are called mentions, a set of
mentions that all refer to the same discourse entity
is called a \revb{coreference }chain.
Annotated corpora are important resources
for developing and evaluating automatic coreference resolution methods.

Turkish is an agglutinative language and
Turkish coreference resolution poses several challenges
different from many other languages,
in particular the absence of grammatical gender,
the possibility of null pronouns in subject and object position,
possessive pronouns that can be expressed as suffixes,
and ambiguities among possessive and number morphemes,
e.g., \quo{çocukları} can be analysed as
\quo{their children} or as \quo{his/her children},
depending on context \cite{Oflazer1994}.

No coreference resolution corpus exists for Turkish so far.
We here describe the result of an effort
to create such a corpus based on the
METU-Sabanci Turkish Treebank
(Say, Zeyrek, Oflazer, and \"{O}zge, 2004;
Atalay, Oflazer, and Say, 2003;
Oflazer, Say, Hakkani-T{\"{u}}r, and T{\"{u}}r, 2003)
which is, to the best of our knowledge,
the only publicly available Turkish Treebank.

Our contributions are as follows.
\bi
\item
  We describe two stages of annotation:
  in \phaseone, annotators created mentions and \revb{coreference }chains,
  which did not yield sufficient inter-annotator agreement.
  In \phasetwo, mentions were given to annotators who created only \revb{coreference }chains.
  We collected on average more than ten independent annotations per document
  for each document in the METU-Sabanci Turkish Treebank.
\item
  We describe annotator profiles and adjudication,
  which was done semi-automatically in \phaseone\ and
  fully automatically in \phasetwo.
  We describe the principles of our automatic adjudication tool
  which uses a voting-like approach.
  Such an automatic approach is possible because we collected
  enough %
  annotations per document.
\item
  We describe the XML format used to address documents,
  sentences, and tokens in the METU-Sabanci Turkish Treebank.
  We provide a public version of the corpus as XML,
  including tools to convert the corpus to CoNLL format.
  (For licensing reasons we cannot re-publish the Turkish Treebank data.)
\item
  We describe and provide a baseline method
  for mention detection
  and coreference resolution, compatible with the format
  of the corpus.
  We evaluate this baseline method on the corpus with
  leave-one-out cross-validation.
\ei

Section~\ref{secPrelimsRelated} gives preliminaries of
coreference resolution and the Turkish language
and describes related work.
Section~\ref{secCorpus}
explains the annotation and adjudication process
and discusses properties of the corpus,
annotator profiles,
and supporting tools.
Section~\ref{secBaseline} describes the baseline system
and its evaluation on the corpus.
Section~\ref{secConclusion} concludes and gives an outlook
on future work.

We provide the Marmara Turkish Coreference Corpus,
tools, and the baseline system, in the following public repository:
\mbox{{\tt\small{}https://bitbucket.org/knowlp/marmara-turkish-coreference-corpus}\,}.

\section{Preliminaries and Related Work}
\label{secPrelimsRelated}

Next, we give background information and related work on coreference resolution,
the Turkish language, and specific challenges
of coreference resolution in Turkish.

\subsection{Coreference Resolution}
Coreference resolution is the task of marking phrases
that refer to the same discourse entity as coreferent.
Mention detection, which identifies such phrases,
is usually included in that task.
Coreference resolution is not limited to resolving pronouns:
in \revb{computational }linguistics it was first introduced as a benchmark
for \emph{deep semantic understanding} of text in the
message understanding conference \revb{(MUC) }series \cite{Grishman1995muc6,Hirschman1998muc7coref}.
After MUC, the Automatic Content Extraction (ACE)
program
(Doddington, Mitchell, Przybocki, Ramshaw, Strassel, and Weischedel, 2004)
required English coreference resolution
as foundation for all tasks of years 2000--2004.
The SemEval competition series followed ACE
and featured the first multilingual coreference
resolution challenge in 2010~%
(Recasens, M{\`{a}}rquez, Sapena, Mart{\'{i}}, Taul{\'{e}}, Hoste, Poesio, and Versley, 2010).
The %
freely available, large, and multilingual
OntoNotes corpus
(Hovy, Marcus, Palmer, Ramshaw, and Weischedel, 2006)
was used in the multilingual coreference task in CoNLL-2012~%
(Pradhan, Moschitti, Xue, Uryupina, and Zhang, 2012)
and contains coreference annotations for English, Arabic, and Mandarin
(Pradhan, Ramshaw, Weischedel, MacBride, and Micciulla, 2007).
{\revbmark
The above mentioned corpora differ with respect to their coreference annotation principles.
MUC and ACE corpora include only noun phrases while OntoNotes also includes heads of verb phrases (and elided subjects/objects for Chinese and Arabic).
The ACE corpus includes only certain types of (military relevant) entities.
The ACE corpus includes singleton mentions while the MUC and OntoNotes corpora do not include singletons.
Predication is annotated only in the ACE corpus,
without discriminating it from identity coreference.
Appositions are annotated in all three corpora;
however, only in OntoNotes the annotation distinguishes
apposition from identity coreference.
The MUC corpus, moreover, includes for each mention a minimal sub-span that is relevant for scoring overlapping mentions.
For more details about these corpora and their differences,
we refer to Poesio, Pradhan, Recasens, Rodriguez, and Versley~\shortcite{Poesio2016annotatedcorpora}.}

Coreference resolution has been surveyed
by Ng~\shortcite{Ng2010coref15}.
Approaches are manifold
and \revb{based on unsupervised and supervised machine learning methods},
rule-based systems, and combinations.
\revb{An example for an unsupervised noun phrase coreference resolution approach
based on clustering is the work of Cardie and Wagstaff~\shortcite{Cardie1999corefcluster}.}
In most \revb{supervised }approaches,
equivalence relations of \revb{coreference }chains
are assembled from
predictions of the relatedness of pairs of mentions.
\revb{An early }machine learning approach \revb{of that kind}
is due to
Soon, Ng, and Lim~\shortcite{Soon2001},
methods for building \revb{coreference }chains from link predictions
include local greedy heuristics
as done by
Bengtson and Roth~\shortcite{Bengtson2008}
or Stoyanov and Eisner~\shortcite{Stoyanov2012},
global optimization formulations
such as relaxation labelling
(Sapena, Padro, and Turmo, 2012)
or ranking with ILP or Markov Logic~%
(Culotta, Wick, and McCallum, 2007;
Denis and Baldridge, 2009)
and representations of trees of links~%
(Fernandes, dos Santos, and Milidi{\'{u}}, 2012;
Chang, Samdani, and Roth, 2013).
The first rule-based algorithm for anaphora resolution
was done by Hobbs~\shortcite{Hobbs1978}.
More recent rule-based systems merge \revb{coreference }chains based on
several sets of rules in a multi-stage filtering approach~%
(Lee, Chang, Peirsman, Chambers, Surdeanu, and Jurafsky, 2013);
moreover, there are hybrid systems combining rules and machine learning such as the one by Chen and Ng~\shortcite{Chen2012conll}.
Other approaches use curated or distributed knowledge sources
such as WordNet, Google distance, and Wikipedia~%
(Poesio, Mehta, Maroudas, and Hitzeman, 2004;
Zheng, Vilnis, Singh, Choi, and McCallum, 2013).

\revb{%
Recently, several coreference resolution approaches
based on word embeddings were introduced.
Word embeddings
are vector representations of words
that are learned in an unsupervised way from an text corpus.
Embedding vectors are motivated by the idea that a word should be known
by the company it keeps.
These vectors are learned with the goal of making them
similar if the respective words occur in similar contexts
(for example if they co-occur with similar words in a neighbourhood of limited distance).
Embeddings vectors capture semantic properties of words
and have been shown to be useful for many NLP tasks.
Prominent word embedding approaches are
word2vec (Mikolov, Sutskever, Chen, Corrado, and Dean, 2013),
\nocite{Mikolov2013word2vec}%
GloVe (Pennington, Socher, and Manning, 2014),
\nocite{Pennington2014glove}%
and FastText (Bojanowski, Grave, Joulin, and Mikolov, 2017).
\nocite{Bojanowski2017fasttext}
Coreference resolution approaches based on word vectors
are often based on neural networks,
for example those
by Lee, He, Lewis, and Zettlemoyer~\shortcite{Lee2017neuralcoref}
and by Wu and Ma~\shortcite{Wu2017deepcoref},
but there are also approaches based on Support Vector Machines (SVM) \cite{Cortes1995svm}
such as the one by \cite{Simova2017embeddingscoref}.
Importantly, these methods
do not require preprocessing with a parser or named entity recognizer,
although Wu et al.~structure the neural network into components
that are reminiscent of parsing and named entity recognition modules.
}%

Note, that anaphora resolution \cite{Hirst1981,Mitkov2002}
is a problem orthogonal to coreference resolution
\cite{VanDeemter2000},
because anaphora resolution focuses on referring expressions
that point to previous expressions in the text.
Cataphora (i.e., pronouns pointing to later occurrences in the text)
are excluded.
On the other hand, different from most works on coreference,
anaphora resolution includes bound pronouns
that do not refer to concrete entities
because they are quantified using, e.g., \quo{some} or \quo{none}.

\subsubsection{Coreference Resolution System Evaluation Metrics}
\label{secCorefMetrics}
{\revamark%
We next recall several evaluation metrics that have been
defined for evaluating the output of a system
that predicts mentions and coreference chains
for a given input document.
\revb{Note, that these metrics are suitable for evaluating \emph{systems} only.
For the equally important task of
evaluating the \emph{reliability of human annotators},
inter-annotator agreement metrics exist
(see Section~\ref{secIAAMetrics}).}

Formally, a document $D$ is a sequence of tokens $D = t_1,\ldots,t_n$,
a mention is a span $(f,t)$ with $1 \les f \les t \les n$ over $D$,
and an entity (also called coreference chain)
is a set of mentions over $D$.
Given a set $K$ of \emph{key} entities
and a set $R$ of \emph{response} entities
over the same document $D$,
an evaluation metric defines a score between 0 and 1
over $K$ and $R$.

The coreference scoring metrics used in the reference
coreference scorer
(Pradhan, Luo, Recasens, Hovy, Ng, and Strube, 2014)
and used in our evaluation are
MUC by Vilain, Burger, Aberdeen, and Connolly~%
\shortcite{Vilain1995muc6coref},
B$^3$ by Bagga and Baldwin~\shortcite{Bagga1998},
CEAF$_m$ and CEAF$_e$ by Luo~\shortcite{Luo2005},
and BLANC by Recasens and Hovy~\shortcite{Recasens2010blanc}.
These metrics have in common
that partially overlapping mentions
and non-overlapping mentions are treated the same:
two mentions are either counted as equal
or as inequal.
For that reason, we describe the above mentioned metrics
in the following simplified manner:
we leave the document $D$ unspecified,
we consider a set $K$ of key entities
and a set $R$ of response entities,
and we let the set $M$ of mentions be defined implicitly
as $M = \bigcup K \cup \bigcup R$.
We follow
Pradhan et al.~\shortcite{Pradhan2014scorer}
and
Sapena, Padr\'{o}, and Turmo \shortcite{Sapena2008corefsurvey}
for the following description of metrics
and denote by $K_1,\ldots,K_{n_k}$ the entities in $K$,
and by $R_1,\ldots,R_{n_r}$ the entities in $R$.
{\revbmark
A \emph{link} is a pair $(m_1,m_2)$
of distinct mentions, $m_1, m_2 \ins M$.
If the mentions are in the same coreference chain,
the link is called a \emph{coreference} link,
otherwise it is called a \emph{non-coreference link}.}

MUC
(Vilain, Burger, Aberdeen, and Connolly, 1995)
is a \emph{link-based metric}
based on the \revb{minimum }number of links between mentions
that are required for defining an entity.
MUC Recall and Precision are defined as
\begin{align*}
  R &= \frac{
      \sum_{i=1}^{n_k} \left( |K_i| - |p(K_i)| \right)
    }{
      \sum_{i=1}^{n_k} \left( |K_i| - 1 \right)
    }
  &&\text{and}
  &
  P &= \frac{
      \sum_{j=1}^{n_r} \left( |R_j| - |p'(R_j)| \right)
    }{
      \sum_{i=1}^{n_r} \left( |R_j| - 1 \right)
    }
\end{align*}
where $p(K_i)$ is the set of partitions
created by intersecting $K_i$ with response entities,
and, \revb{conversely, }$p'(R_j)$ is the set of partitions
created by intersecting $R_j$ with key entities.
F1-score is defined as
the harmonic mean of Precision and Recall.

MUC gives the same score reduction for \revb{incorrectly }merging two big coreference chains
and for \revb{incorrectly placing one mention into the wrong coreference chain},
which is counterintuitive.
Because it is link-based,
MUC \revb{cannot handle singleton mentions}.
The B$^3$ metric aims to overcome these drawbacks
by giving a score based on \emph{mentions}.

The B$^3$ metric~\cite{Bagga1998}
defines Precision and Recall
for each key mention and accumulates a score over the whole document.
B$^3$ Recall and Precision are defined as
\begin{align*}
  R &= \frac{
      \sum_{i=1}^{n_k} \sum_{j=1}^{n_r}
        \frac{|K_i \caps R_j|^2}{|K_i|}
    }{
      \sum_{i=1}^{n_k} |K_i|
    }
  &&\text{and}
  &
  P &= \frac{
      \sum_{j=1}^{n_r} \sum_{i=1}^{n_k}
        \frac{|K_i \caps R_j|^2}{|R_j|}
    }{
      \sum_{i=1}^{n_r} |R_j|
    }.
\end{align*}
\revb{Because the B$^3$ metric intersects key and response entities,
one mention can contribute to the score multiple times,
leading to counterintuitive scores.
To overcome this limitation,
the CEAF metrics were proposed.}

The family of \revb{Constrainted Entity-Alignment F-Measures (CEAF)} by
Luo~\shortcite{Luo2005}
is centred around \emph{entities}:
given a similarity measure
$\Phi : M \times M \to \mathbb{R}$
that determines how \revb{well }two entities match,
CEAF first finds the best \emph{one-to-one} mapping
$g^\star : \{1,\ldots,n_k\} \to \{1,\ldots,n_r\}$
between key and response entity indexes,
i.e., the mapping such that
$\sum_{(i,j) \in g^\star} \Phi(K_i,R_j)$
becomes maximal among all possible mappings.
Because of this mapping,
\revb{each key and each response mention
contributes exactly once to the overall CEAF score,}
which \revb{produces more realistic scores than }MUC and B$^3$.
CEAF Recall and Precision are defined as
\begin{align*}
  R &= \frac{
      \sum_{(i,j) \in g^\star} \Phi_\alpha(K_i,R_j)
    }{
      \sum_{i \in \{1,\ldots,n_k\}} \Phi_\alpha(K_i,K_i)
    }
  &&
  \text{and}
  &
  P &= \frac{
      \sum_{(i,j) \in g^\star} \Phi_\alpha(K_i,R_j)
    }{
      \sum_{j \in \{1,\ldots,n_r\}} \Phi_\alpha(R_j,R_j)
    }
\end{align*}
where $\alpha \in \{ m, e \}$
specifies one of two metrics:
CEAF$_m$ computes entity-entity similarity
according to the size of the intersection of entities,
formally $\Phi_m(K_i,R_j) = |K_i \caps R_j|$;
CEAF$_e$ normalizes this similarity
according to the size of both entities,
formally $\Phi_e(K_i,R_j) = \frac{2|K_i \caps R_j|}{|K_i|+|R_j|}$.

\revb{Both the B$^3$ and the CEAF metrics
evaluate an assignment of a mention to a coreference chain
independent from the size of the chain.
To overcome this limitation, the BLANC metric was proposed.}

\newcommand{\mylinks}[1]{\mi{\revbmark{}crlinks}(#1)}
The \revb{BiLateral Assessment of Noun-phrase Coreference (BLANC)} metric \cite{Recasens2010blanc}
gives equal importance to \revb{coreference }links
and \revb{non-coreference links}.
The motivation for creating BLANC was to correctly handle
singleton mentions and to handle \revb{coreference }chains with many mentions
and with few mentions more fairly than possible with B$^3$ and CEAF.
We here show the BLANC extension by
Luo, Pradhan, Recasens, and Hovy \shortcite{Luo2014blanc2}
which is able to process spurious as well as missing response mentions.
Given a set of entities $X$
we denote by $\mylinks{X}$
the set $\{ (m_1,m_2) \mids m_1, m_2 \in Y, Y \in X, m_1 \neq m_2 \}$
of all \revb{coreference }links \revb{that define }entities in $X$.
We denote by $C_k = \mylinks{K}$ and $C_r = \mylinks{R}$
the set of coreference links in key and response;
moreover, by $T_k = \mylinks{\{\bigcup K\}}$ and $T_r = \mylinks{\{\bigcup R\}}$
the set of all possible key and response links;
and finally, by $N_k = T_k \setminus C_k$ and $N_r = T_r \setminus C_r$
the sets of non-coreference links in key and response, respectively.
Recall, Precision, and F-score of coreference links
are defined as
\begin{align*}
  R_c &= \frac{|C_k \caps C_r|}%
    {|C_k|}\text{ , }
  &
  P_c &= \frac{|C_k \caps C_r|}%
    {|C_r|}\text{ , and}
  &
  F_c &= \frac{2 R_c P_c}{R_c + P_c}\text{ ,}
\intertext{
and the same metrics are also defined
for non-coreference links:
}
  R_n &= \frac{|N_k \caps N_r|}%
    {|N_k|}\text{ , }
  &
  P_n &= \frac{|N_k \caps N_r|}%
    {|N_r|}\text{ , and}
  &
  F_n &= \frac{2 R_n P_n}{R_n + P_n}\text{ .}
\end{align*}
Finally,
BLANC is the arithmetic mean of F-measure of
coreference and non-coreference \revb{links},
that is $\mi{BLANC} = \frac{F_c+F_n}{2}$.}

{\revbmark
Moosavi and Strube \shortcite{Moosavi2016leametric}
propose the Link-based Entity-Aware (LEA) metric which overcomes
the \emph{mention identification effect}
of the B$^3$, CEAF, and BLANC metrics:
adding incorrect entities to the system output decreases Recall
of these metrics which makes them no longer reliable.
The LEA metric scores each coreference chain according to
its importance (in terms of its size)
and according to how well it is resolved
(in terms of coreference links reproduced in the response).
Given a coreference chain $C \subseteqs M$,
the number of links in $C$ is
$\mi{link}(C) \eqs \frac{|C|(|C|\,{-}\,1)}{2}$
and LEA Recall and Precision are defined as
\begin{align*}
  \mi{R_{\mi{LEA}}} &= \frac{
      \Sigma_{i \eqs 1}^{n_k} \left( |K_i| \cdot
        \Sigma_{j \eqs 1}^{n_r} \frac{\mi{link}(K_i \cups R_j)}{\mi{link}(K_i)}
      \right)
    }{
      \Sigma_{z \eqs 1}^{n_k} |K_z|
    }\text{ and} \\
  \mi{P_{\mi{LEA}}} &= \frac{
      \Sigma_{i \eqs 1}^{n_r} \left( |R_i| \cdot
        \Sigma_{j \eqs 1}^{n_k} \frac{\mi{link}(R_i \cups K_j)}{\mi{link}(R_i)}
      \right)
    }{
      \Sigma_{z \eqs 1}^{n_r} |R_z|
    }\text{ .}
\end{align*}
LEA F1-score is computed as the harmonic mean of LEA Precision and Recall.
}

\subsubsection{Coreference Inter-Annotator Agreement Metrics}
\label{secIAAMetrics}
{\revamark%
Inter-annotator agreement metrics are a tool
for quantifying the reliability of human annotators
who independently annotated the same document.
Different from system evaluation metrics,
inter-annotator agreement is computed without a gold standard
and for (potentially) more than two annotations at once.
\revb{For a detailed survey and justification of inter-annotator metrics,
in particular their difference to system evaluation metrics (see Section~\ref{secCorefMetrics}),
we refer to Artstein and Poesio~\shortcite{Artstein2008agreementmetrics}.}

Krippendorff~\shortcite{Krippendorff1980}
defined metric $\alpha$ for quantifying the reliability of
$r$ classification decisions of $m$ annotators:
\begin{align*}
  \alpha &= 1 - \frac{rm-1}{m} \frac{
    \sum_i \sum_b \sum_{c > b} n_{b_i} n_{c_i} \delta_{bc}
  }{
    \sum_b \sum_c n_b n_c \delta_{bc}
  }
\end{align*}
where $i$ ranges over objects to be classified,
$b$ and $c$ range over classes,
\revb{$n_x$ is the number of %
objects that were put into class $x$ by annotators,}
$n_{x_i}$ is the number of times object $i$ was put into class $x$ by annotators,
and $\delta_{bc}$ is a distance function between classes $b$ and $c$.
An annotation process is considered reliable
if $\alpha > 0.67$.

When applying this metric to coreference annotation,
we consider mentions as objects,
and entities that were produced by annotators as classes.
It is useful to create a fine-grained distance function $\delta$
between entities,
for example putting mention $A$ into entity $E_1 = \{A,B,C,D\}$,
putting it into entity $E_2 = \{A,C,D\}$, and putting it into entity $E_3 = \{A,E\}$
intuitively is a mistake of varying severity.
In this work, we use the following coreference-specific variations of $\alpha$.
We denote by $\mi{IAA}_1$
the metric defined by Passonneau~\shortcite{Passonneau2004}
where $\delta_{bc} = 1 - M_{bc}$
where the match score $M_{bc}$ obtains a value of 1 for equality,
$\frac{2}{3}$ if $b$ is a subset of $c$ or $c$ is a subset of $b$,
$\frac{1}{3}$ if $b$ and $c$ are intersecting in more than a single mention,
and 0 otherwise.
We denote by $\mi{IAA}_2$
the metric defined by Passonneau, Habash, and Rambow~%
\shortcite{Passonneau2006masi}
where $\delta_{bc} = 1 - J_{bc} M_{bc}$,
and $J_{bc}$ is the Jaccard distance between sets $b$ and $c$.
Metric $\mi{IAA}_2$ has the advantage
that it normalizes over heterogeneous sizes of entities.}

\subsection{Turkish}
Turkish is a member of the family of Altaic languages,
it is an agglutinative language where suffixes
are attached to a root word.
Derivational and inflectional suffixes are very productive
(Oflazer, 1993;
Oflazer, G{\"{o}}{\c{c}}men, and Bozşahin, 1994)
\nocite{Oflazer1993turkishmorphologyposter}\nocite{Oflazer1994}%
and are subject to vowel harmony from the root word.
Morphological analysis is challenging
due to ambiguities between different types of suffixes,
for example \quo{izin} can mean
\quo{your trace} (iz+P2Sg+Nom),
\quo{trace} (iz+Pnon+Gen),
or \quo{permission} (izin+Pnon+Nom)
(Hakkani-T{\"{u}}r, Oflazer, and T{\"{u}}r, 2002).

\subsubsection{The Turkish Treebank}
The METU-Sabanci Turkish Treebank
(\revb{hereafter referred to as the }Turkish Treebank)
\cite{Atalay2003,Oflazer2003turkishtreebank}
contains a subset of the METU Turkish Corpus
\cite{Say2004}
in tokenized form.
Each token is analysed morphologically
and split into inflectional groups (IGs).
Sentences are annotated with dependency parse information,
where dependencies point to specific IGs within tokens.
The Turkish Treebank splits tokens
into IGs on derivational boundaries,
for example, \quo{evimdekiler}
(those in my house)
is analysed \cite{Oflazer2003turkishtreebank} as
\smallskip

\centerline{ev+Noun+A3sg+P1sg+Loc\textasciicircum{}DB+Adj\textasciicircum{}DB+Noun+Zero+A3pl+Pnon+Nom}
\smallskip

\noindent
where \textasciicircum{}DB indicates derivation boundaries and the token consists of three IGs
\quo{evimde} (in my house), \quo{ki} (adjectivization),
and \quo{ler} (nominalization+plural).
A CoNLL format that provides a CoNLL token
corresponding to each IG of a token has been created
for Turkish dependency parsing~\cite{Buchholz2006}.

Named entities in the Turkish Treebank are not marked specially,
but multiword named entities are represented as single tokens.

\subsubsection{Turkish Coreference Resolution}
\label{secTurkishCoref}
The following properties of Turkish \revb{and the Turkish Treebank}\
are particularly relevant for coreference resolution.

\emph{\revbmark{Accessibility of morphemes as markables.}}
In the above example,
\quo{those in my house} as well as \quo{my house} as well as
\quo{my} could be coreferent with mentions in the document.
However, neither \quo{my house} nor \quo{my} is available
as a separate unit of analysis: both are parts of the first IG (\quo{evimde}).

\emph{\revbmark{}Gender.}
Gender is not marked in Turkish
with the exception of the honorifics \quo{Bey} and \quo{Hanım}
which corresponds to English \quo{Mr} and \quo{Mrs}.
Moreover, several common first names
apply to both genders.
Hence, gender-based syntactic compatibility checks
for mentions are only possible in some cases.

\emph{\revbmark{}Personal pronoun subjects.}
In Turkish, these subjects are usually
realized as suffixes of the verb,
e.g., \quo{gidiyoruz} (we are going)
and \quo{gidiyorlar} (they are going)
but they can also be realized explicitly
as in \quo{biz gidiyoruz},
depending on discourse conditions \cite{Turan1996}.

\emph{\revbmark{}Proper noun suffixes.}
Suffixes of proper nouns in written Turkish
are systematically separated from the proper nouns
using a single quote,
e.g., \quo{Türkiye'den} (from Turkey)
and \quo{Türkiye'deki} (the thing in Turkey).
This rule simplifies the finding of
equal proper noun mentions in
coreference resolution for Turkish.

Most works about referring expressions in Turkish
focus on anaphora resolution and not on full coreference resolution.
One exception is the work of
Küçük and Yazıcı~\shortcite{Kucuk2008videocoref}
on political news texts extracted from videos:
they focus on Gazetteers for extracting mentions
(without considering general NPs or syntactic information),
provide a rule-based heuristic
based on recency for creating coreference chains,
and evaluate their approach on three documents
(which are not part of the Turkish Treebank).

\subsubsection{Turkish Anaphora Resolution}
\revb{%
Next, we describe work on Turkish anaphora resolution
which is related to coreference resolution.}

Erkan and Akman~\shortcite{Erkan1998}
describe an implementation of \revb{pronominal }anaphora resolution
in a framework for situation theory
which is based on knowledge representation
and logical reasoning.
Hobb's na\"{\i}ve pronoun resolution algorithm~%
\cite{Hobbs1978}
was realized for Turkish and tested
on 10 toy sentences~%
\cite{Tufekci2007}.

Centering theory~%
(Grosz, Joshi, and Weinstein, 1995)
is the foundation
of several works on Turkish pronouns.
Turan~\shortcite{Turan1996} performed
a study about discourse conditions for referring vs.~nonreferring expressions and null vs.~overt pronouns,
and evaluated the theory on 2500 annotated tokens.
Yüksel and Bozşahin~\shortcite{Yuksel2002}
created a system for generating referring expressions that was tested on a machine translation task.
Furthermore,
there is a theoretical model of anaphora resolution
based on Centering Theory
by Yıldırım, Kılı{\c{c}}aslan, and Ayka{\c{c}}~\shortcite{Yildirim2004}.

Küçük and Yöndem~\shortcite{Kucuk2007}
described a system for finding and resolving Turkish
pronominal anaphora and annotated
12266 anaphora candidate instances
in the METU Turkish Corpus to evaluate their candidate extractor and decision tree learner for anaphora resolution.
Kılıçaslan, G{\"{u}}ner, and Yıldırım~%
\shortcite{Kilicaslan2009} performed
a comprehensive study on pronoun resolution
and evaluated various machine learning methods
for resolving overt and null pronouns
in a corpus of 20 \revb{stories for }children.
\section{Marmara Turkish Coreference Corpus}
\label{secCorpus}

We next describe the annotation and adjudication process
including formal adjudication criteria,
key properties of the resulting corpus,
and supporting tools.

\subsection{Annotation Process}
\label{secProcess}

Annotations were collected from computer engineering
students participating in a lecture on natural language processing,
after educating them in basic linguistic analysis and coreference resolution.
To achieve reasonable annotation quality,
we aimed to keep the annotation principles simple and
therefore based them on few rules and examples.
We designed an annotation manual~\cite{Surmeli2015annotationmanualv1}
{\revamark%
and a revised version of this manual
(S{\"{u}}rmeli, Cıngıllı, Tun{\c{c}}er, and Sch{\"{u}}ller, 2016)}
for marking coreference according to the following principles:
\begin{itemize}
\item
  all {\revamark specific} entities that are mentioned more than once
  \revb{by a noun phrase, pronoun, or nominalized adjective, }shall be annotated,
\item
  mentions shall be marked as the biggest possible
  span of tokens that describes the entity,
\item
  lists shall not be annotated (elements of lists can be annotated), and
\item
  predications shall not be annotated.
\end{itemize}
By marking mentions as the biggest \revb{possible }spans,
\revb{a phrase and potentially existing appositive phrases become part of the same mention}.
\revb{This }is different from OntoNotes
where \revb{phrases and }appositives are \revb{separate mentions
which are put into }a special type of \revb{appositive }coreference \revb{chain}.
We do not mark predications because they are a different
type of coreference as argued by van Deemter and Kibble~\shortcite{VanDeemter2000}.
{\revamark%
Figure~\ref{figExampleOmurUzatma} shows an example of Turkish mentions
and coreference chains.}
\myfigureExampleOmurUzatma

{\revamark%
Specific entities introduce or refer to a specific discourse entity
while non-specific entities are variables over sets of potential discourse entities.
Non-specific entities are usually indicated by quantifier words
such as ``everybody'' or ``some''.
Figure~\ref{figExampleEv} shows an example from the annotation manual
\cite{Surmeli2016annotationmanualv2}
where ``Boş, kiralık apartman dairesi''
(an empty apartment flat that is for rent)
and ``o'' (it) has an anaphoric relationship
but we cannot pinpoint a specific empty flat;
therefore, no coreference shall be annotated.}

Figure~\ref{figProcess} visualizes the process that led to the final corpus.
Annotations were collected in two phases:
\phaseone\ took place in October--December 2015 and \phasetwo\ during October--December 2016.
{\revamark%
\phaseone\ used the initial version of the annotation manual \cite{Surmeli2015annotationmanualv1} and \phasetwo\ the revised version \cite{Surmeli2016annotationmanualv2}.
The final corpus resulting from this project contains
coreference annotations only from \phasetwo.}
\myfigureExampleEv
\myfigureprocess
In \phaseone, annotations were created by 19 annotators
with the
\revb{\quo{Co-reference Editor} that is part of the \quo{Language Resources} functionality
of }GATE\revb{ Developer}~%
(Gaizauskas, Cunningham, Wilks, Rodgers, and Humphreys, 1996;
Cunningham, Tablan, Roberts, and Bontcheva, 2013).
\revb{%
We preferred GATE because it provided
an XML interface (which was compatible with the Turkish Treebank format),
a well-structured website and documentation,
and a comfortable installation procedure that works on multiple platforms.}
\revb{\phaseone\ }yielded on average 6.5 annotations per document
for 21 documents in the Treebank.
Adjudication of these documents was done semi-automatically
(see Sections~\ref{secAdjudication} and~\ref{secTools}).
However, due to low inter-annotator agreement
about mention boundaries,
decisions often depended on the adjudicator.
\revb{Each unique annotated mention in \phaseone\ was annotated by only~1.9 annotators on average,
where perfect agreement would be 6.5, i.e., the number of annotators per document.}
Moreover, we identified several issues in the annotation manual.
{\revamark%
Therefore,
we created a revised version \cite{Surmeli2016annotationmanualv2}
of the annotation manual which included additional examples\revb{,
in particular about the difference between specific and non-specific mentions,
and about the difference between coreference and predication}.}
Moreover, in order to make the setting simpler,
we decided to perform a second annotation phase
where we collect annotations with \emph{given} mentions.
We used the list of mentions resulting from the adjudicated documents of \phaseone.
Mentions for those 12 documents that were not annotated in \phaseone\
were manually created in a collaboration of two annotators for each document.
{\revamark%
Mentions were annotated whenever there was doubt about them.
Therefore, they are available for coreference annotation
but can be omitted if no coreferent mention exists.
This coincides with the strategy used in mention prediction systems
which usually aim for high Recall and leave the elimination of spurious mentions
to the coreference resolution prediction system.}

In \phasetwo, 46 annotators were given CoNLL files
with token and coreference columns
where each mention was given in its own \revb{coreference }chain.
Annotators created \revb{339 individual annotation files }%
with equalities between \revb{coreference }chain IDs
and uploaded these files to a web service
where they were checked for syntactical correctness.
The submission file format is described by
S{\"{u}}rmeli et al. \shortcite{Surmeli2016annotationmanualv2}.
This method of collecting annotation as text files
might seem archaic; however, in practice,
annotators were more comfortable with such a system
than with the graphical user interface of GATE in \phaseone.
We were not able to use the BRAT~%
(Stenetorp, Pyysalo, Topic, Ohta, Ananiadou, and Tsujii, 2012)
annotation tool
because of difficulties
representing sentence and word addresses
in a way that they can be extracted from annotations.

\revb{The problem of disagreement on mention boundaries was successfully prevented in \phasetwo:
each unique mention was annotated as part of a coreference chain by 9.6 annotators on average,
where perfect agreement would be 10.3.
Therefore, }\phasetwo\ yielded
of sufficient inter-annotator agreement
to perform fully automatic adjudication (see next section).

\paragraph{Annotator Profiles.}
Anonymized learner profiles were collected from all students
in \phasetwo\ (written permission for using and publishing the data was also obtained).
Annotators are on average 23 years old university students
at Marmara University in Istanbul.
Out of 46 annotators, 29 are male and 17 are female.
One annotator indicated Azerbaijani as a native language,
all others indicated Turkish as one of their native languages.
(Azerbaijani is close to Turkish.)
Two annotators indicated Kurdish as further native language,
and one each Arabic, English, and Macedonian.
Primary and secondary school education was Turkish for 43 annotators, English for two and Azerbaijani for one.
Moreover, 43 annotators lived at least 20 years in predominantly Turkish-speaking communities,
the remaining annotators answered 4, 5, and 14 years, respectively, to this question.
According to this data we consider our annotators to be
capable of understanding and annotating the texts in the corpus
\revb{on the level of or close to the level of a native speaker}.

\subsection{\revb{Analysis of Annotations}}
\label{secAnnotationproperties}
\myannotationtable
Table~\ref{tblAnnotationMetrics} shows properties
\revb{of the annotations that were collected.}

Over all documents,
\revb{the inter-annotator agreement metric}
$\mi{IAA}_1$ is 76\% and $\mi{IAA}_2$ is 90\%,
which indicates reliability of our annotation process
{\revamark (see Section~\ref{secIAAMetrics})}.
We observe worse IAA for genres that are focused on writing as an art form,
i.e., for genres Short Story and Other (a first-person narrative).
\revb{These genres contain a high percentage of dialogues
with frequent changes between speaker and addressee,
which led to a higher number of annotator mistakes regarding personal pronouns.}

By comparing column $\mi{GM}$ and $\mi{AM}$ (\revb{given and }annotated mentions\revb{, respectively})
we see that annotators rarely use all mentions in the annotated \revb{coreference }chains.
Annotators were instructed to omit mentions from annotation
if there was no other specific mention referring to exactly the same discourse entity.
To reduce the chance that these mentions were omitted due to an oversight,
the annotation submission system indicated which mentions were left unused.
{\revamark%
Very few annotators asked for additional mentions
(and only in a single case, a mention was actually missing).
In summary, the difference between $\mi{GM}$ and $\mi{AM}$ indicates \revb{that}
our coreference annotators consciously omitted certain mentions from annotation.
This coincides with the strategy of annotating mentions with high Recall,
and relying on coreference annotators for obtaining high Precision of mentions.}

Column $\mi{Ph1}$ indicates how many of the documents
were annotated in both phases of the annotation process.
For example, the News genre contains 9 documents.
Mentions of 2 News documents were obtained from \phaseone,
the others from \phasetwo\ (see also Figure~\ref{figProcess}).

\subsection{\revb{Semi-automatic} Adjudication}
\label{secAdjudication}
We collected an average of 10.3 annotations per document
\revb{(see Table~\ref{tblAnnotationMetrics})}.
This high number of annotations,
combined with the observed IAA,
allows us to automatically adjudicate the corpus.
This is different from other coreference annotations,
in particular from OntoNotes where two annotators
created annotations followed by adjudication
done by a single human expert~%
(Weischedel, Pradhan, Ramshaw, Kaufman, Franchini, El-Bachouti, Xue, Palmer, Hwang,
Bonial, Choi, Mansouri, Foster, Hawwary, Marcus, Taylor, Greenberg, Hovy, Belvin,
and Houston, 2012).
Our automatic adjudication method is based on combinatorial optimization:
we search for a solution of \revb{coreference }chains that has overall minimal
divergence from all annotator inputs.
Divergence is measured in terms of links
given and omitted by annotators.

Formally, given a set $M$ of mentions in a document,
{\revamark%
annotators produce $k$ sets of entities
$A_1$, \ldots, $A_k$ over $M$,
that is each $A_i$, $1 < i < k$, contains a partition (disjoint subsets) of $M$.
A solution $G$ also contains a partition of $M$,}
and we search for $G$ such that the following objective becomes minimal:
\begin{equation}
  \label{eqObj}
  \sum_{\substack{
    m, m' \ins M \\
    i \ins \{1, \ldots, k\}}}
      \big(
      2 \cdot a\left(m,m',A_i\right)\cdot na\left(m,m',G\right)
      + na\left(m,m',A_i\right)\cdot a\left(m,m',G\right)
      \big)
\end{equation}
where $a(m,m',A)$ indicates \revb{whether }%
$m$ and $m'$ are coreferent in $A$,
formally
\begin{equation*}
  a(m,m',A) = \left\{\begin{array}{ll}
    1 & \text{if $\exists C \in A: \{m,m'\} \subseteq C$} \\
    0 & \text{otherwise.}
  \end{array}\right.
\end{equation*}
{\revamark%
Similarly, $na(m,m',A)$ indicates \revb{whether }%
$m$ and $m'$ are not coreferent in $A$:
formally, $na(m,m',A) = 1 - a(m,m',A)$.
The left term of the sum in \eqref{eqObj} incurs a cost of $2 j$
for each \revb{link $(m,m')$ }%
that \revb{is non-coreferent }in the solution~$G$
contrary to the opinion of~$j$ annotators who annotated it as coreferent.
The right term of the sum incurs a cost of $l$
for each \revb{link $(m,m')$ }%
that \revb{is coreferent }in the solution~$G$
contrary to the opinion of~$l$ annotators who \revb{annotated it as non-coreferent}.
\revb{%
We additionally enforce the following hard constraints:
(i) mentions that overlap cannot be coreferent in the solution, and
(ii) the solution can only contain coreference links
that have been annotated by at least one annotator.
Constraint (i) is specific to our corpus
where we ask annotators to annotate the largest span of a mention,
constraint (ii) prevents the invention of coreference links
due to the merging of coreference chains
and is motivated by observations we made while inspecting annotations.}

Intuitively, the optimization criterion
is based on \revb{coreference and noncoreference }links,
similar to the BLANC evaluation metric \cite{Recasens2010blanc}.
Optimal solutions ignore as little as possible
information from annotators, where putting the mentions into the same entity
as well as not putting them into the same entity is used as information.
\revb{A link that is indicated as coreferent by an annotator but is adjudicated as non-coreferent }in the solution $G$ incurs twice the cost of \revb{a link that was annotated as non-coreferent and is coreferent in }$G$.}
\revb{We introduced this preference into the objective
because we made the following observation:
if fewer than half of the annotators put a mention into the same coreference chain
and the remaining annotators did not annotate the mention as coreferent with any other mention
then the annotation of the minority was correct according to our judgement:
the mention should have been annotated as part of the coreference chain.
From that observation, we concluded
that assigning a mention is more likely to be done intentionally
than omitting a mention from a coreference chain,
and this is reflected in the higher weight
of coreference links compared with non-coreference links in the objective function.}
{\revamark%
As an example, if we obtain entities $\{\{A,B\},$ $\{C,D\}\}$ from \revb{4 }annotators,
\revb{$\{\{A,B\}\}$} from \revb{3 }annotators, and $\{\{C,D,E\}\}$ from \revb{2 }annotators,
the optimal solution is $\{\{A,B\},\{C,D\}\}$\revb{:
coreference links $(A,B)$ and $(C,D)$ in this solution
were annotated as non-coreferent by 2 and 3 annotators, respectively,
which incurs a cost of $2\pluss 3 \eqs 5$;
non-coreference links $(C,E)$ and $(D,E)$ in this solution
were annotated as coreferent by 2 annotators,
which incurs a cost of $2\cdot(2\pluss 2) \eqs 8$.
Therefore, the cost of this solution is $5 \pluss 8 \eqs 13$.}
A link that is \revb{coreferent (respectively non-coreferent)} in all annotations and in the solution
does not incur any cost.
}

\revb{%
We inspected the adjudication result after automatic adjudication
to validate the results of automatic adjudication,
and we analyzed annotator mistakes (see Section~\ref{secAnnotatorMistakes}).
Our adjudication tool (see Section~\ref{secTools})
permits a manual specification of partial coreference chains;
however, performing such a manual override was not necessary,
as we did not encounter mistakes of the automatic adjudication method.
Additional details about the adjudication method and tool
have been described by our group \cite{Schuller2018adjudication}.}

\mycorpustable
\subsection{Corpus Properties}
\label{properties}

\revb{Table~\ref{tblCorpusMetrics} shows key properties
of the adjudicated gold standard.
The corpus contains 5170 mentions and 944 coreference chains.}

\emph{\revb{Mentions.}}
\revb{The a}verage number of tokens and mentions per genre
varies a lot. In particular, the Essay genre
contains texts discussing abstract concepts
like \quo{home} \revb{(see Figure~\ref{figExampleEv}) }and \quo{science} which are not annotated;
{\revamark
therefore, the average number of mentions \revb{per document ($M=89$)} %
is significantly lower than in other genres.
The genre Other contains \revb{a first-person }narrative
\revb{which repeatedly mentions }many person names;
therefore, the number of mention\revb{s (298)} is higher than in other genres.}
{\revbmark%
Figure~\ref{figMentiontypes} shows the distribution of mention types in the adjudicated corpus.
Mentions comprising a single token account for~76\% of all mentions,
with an equal distribution between pronouns, proper nouns, and other single-token mentions
such as \quo{babam} (\quo{my father}).
Figure~\ref{figMultiTokenMentionSizes} shows the distribution of mention length
for the remaining~24\% of mentions which span more than one token:
the majority of these mentions contain just two mentions,
for example the mention \quo{bu dağlara} (\quo{these mountains+Dat}).
There are a few long mentions such as
\quo{şeker, kahve, un, ayçiçeği yağı ve antibiyotiklerin bu dağlara ulaşmasından önceki durumu}
(\quo{the times before sugar, coffee, flour, sunflower seed oil, and antibiotics reached
[became available to the people that are living in] these mountains}).
Of all mentions, 5.7\%  are a nested (i.e., contained) within another mention,
and no mention is nested within a nested mention.}
\myfigurehistogoldmentions

\emph{\revb{Coreference Chains.}}
{\revbmark%
Figure~\ref{figChainlengths} depicts the distribution of coreference chain lengths
in the gold standard (using a logarithmic scale to make single occurrences visible):
\revb{coreference }chains that connect just two mentions
occur more often (365 times) than longer chains.
\revb{Coreference }chains that connect more than ten mentions are rare in the corpus,
although there are also a few large \revb{coreference }chains.
Among those \revb{coreference }chains that contain ten or more mentions,
seven refer to the writer or to the first-person narrator and contain only pronouns,
while the others refer mainly to persons and contain mainly proper nouns mentions.}
\myfigurehistochainlengths

\subsubsection{\revb{Annotator Mistakes}}
\label{secAnnotatorMistakes}
{\revbmark
Overall, annotators produced 9,723 coreference chains containing 51,525 mentions.
Figure~\ref{figUseOmitStats} depicts the number of annotated coreference links
over the percentage of annotators that annotated
coreference of the same link.
The left side of the histogram mainly shows links
that are non-coreferent in the gold standard (due to adjudication),
while the right side shows links that are coreferent in the gold standard.
(To depict the agreement of annotators independent
 from the number of annotations per document,
the histogram shows percentages:
for a documents with nine annotators, a single annotated coreference link contributes 11.1\%;
while for eleven annotators, a single annotated coreference link contributes 9.1\%.)
Nearly all links that were annotated as coreferent by fewer than 30\% of annotators are non-coreferent in the gold standard,
while nearly all links that were annotated as coreferent by at least 50\% of annotators are coreferent in the gold standard.
Between 30\% and 50\%, some links are coreferent and some are non-coreferent in the gold standard.
Whether a link occurs in the gold standard
depends on the global optimality of the solution
and on the satisfaction of structural constraints
as described in Section~\ref{secAdjudication}.
\myfigurehistomistakes

From 9,723 annotated coreference chains,
74\% directly correspond with a coreference chain in the gold standard,
and Figure~\ref{figMistakeStats} visualizes the annotator mistakes that can be measured over the remaining coreference chains.
For this analysis, we established a one-to-one matching between
annotated and gold standard coreference chains,
based on the greatest link overlap (similar to the CEAF metrics).
We then analysed deviations of annotators from the gold standard
relative to this matching.
The majority of mistakes (57\%) are coreference chains with missing mentions.
From these mistakes,
42\% are missing a single mention and 73\% are missing at most three mentions.
One third of mistakes (34\%) are coreference chains containing mentions
that are in distinct coreference chains in the gold standard.
In 66\% of these mistakes, a single mention belonged to another coreference chain
and in 95\% of these mistakes, at most three mentions belonged to another coreference chain.
A few mistakes are about mentions
that are not part of any coreference chain in the gold standard:
7\% of annotated chains contained such mentions (and no other mistakes),
and 2\% of annotated coreference chains contained only such mentions.
}%

\subsection{Tools}
\label{secTools}

For creating this corpus, we built several tools.

\paragraph{Document Extractor.}
The METU-Sabanci Turkish Treebank contains 1960 text fragments,
distributed over 33 documents. Most documents are split over several XML files;
however, there is also one XML file containing two distinct documents.
We provide a tool for extracting documents from the Turkish Treebank
and store each document in a single XML file.
The Turkish Treebank is licensed in a way
that it cannot be redistributed with the Marmara Turkish Coreference Corpus;
therefore, the tool generates document files
from a directory containing the unpacked Turkish Treebank.
Our tool not only creates one XML file for each document,
it also recodes all data to UTF-8 and fixes problematic (non-encoded)
attributes that are present in the original corpus.

\paragraph{Coreference XML format.}
For representing coreference information,
we created an XML format
that contains pointers to sentence and word IDs into documents
extracted from the Turkish Treebank.
A sample of such an XML file with two mentions and one \revb{coreference }chain is as follows.
\begin{Verbatim}
<coref>
  <mentions>
    <mention fromWordIX="1" id="0" sentenceNo="00016112313.1" toWordIX="1">
      Prof._Dr._Semih_Koray'ın
    </mention>
    <mention fromWordIX="1" id="2" sentenceNo="00016112313.2" toWordIX="1">
      Koray
    </mention>
  </mentions>
  <chains>
    <chain>
      <mention mentionId="0">Prof._Dr._Semih_Koray'ın</mention>
      <mention mentionId="2">Koray</mention>
    </chain>
  </chains>
</coref>
\end{Verbatim}
In this example, \quo{Prof.\_Dr.\_Semih\_Koray'ın}
is a mention with ID 0 \revb{(attribute \texttt{id="0"})}\ containing
the token with index 1 \revb{(attributes \texttt{fromWordIX="1"} and \texttt{toWordIX="1"})}
in sentence \quo{00016112313.1}
\revb{(attribute \texttt{sentenceNo="00016112313.1"})}\
of the document assembled from the Treebank.
Moreover, there is a \revb{coreference }chain containing that mention
and another mention that consists of the first token of sentence \quo{00016112313.2}.

\revb{%
A design goal for this XML format was to stay close to the XML format of the Turkish Treebank.
Therefore, tokens are indexed relative to sentence numbers,
and XML attribute names are equal to attributes names in the Turkish Treebank.%
\footnote{\revb{Tokens are }called \quo{word\revb{s}} in the Treebank XML format.}}
Note, that the text between the mention XML tags is used only for readability purposes;
the information about mention content is fully represented in the XML attributes.

\paragraph{CoNLL $\Leftrightarrow$ XML Converters.}
As the CoNLL reference coreference scorer~\cite{Pradhan2014scorer}
is based on CoNLL format,
we provide tools for converting a document and a coreference
XML file into a CoNLL file (and vice versa).
We use XML to be consistent with the Turkish Treebank
and because the Treebank license prevents redistribution.

\paragraph{(Semi-)automatic coreference adjudication tool.}
Merging several distinct coreference annotations
into a single gold standard is a complex task,
in particular if annotators do not agree on mentions.
To simplify this task, we created a tool that merges multiple annotations
into a single solution
according to objective~\eqref{eqObj} from Section~\ref{secAdjudication}.
{\revamark Technically, this optimization is performed with
the knowledge representation formalism Answer Set Programming
(Brewka, Eiter, and Truszczynski, 2011)
which yields provably optimal solutions to combinatorial problems in reasonable time.
Manual intervention for editing mentions and \revb{coreference }chains
is possible in our tool, details about
the file formats
and about how to represent the adjudication problem in a way that is
efficient enough for practical purposes
are described in a separate paper \cite{Schuller2018adjudication}.}
In \phasetwo\ we performed only automatic
adjudication and did not need manual intervention.

For the purpose of this project,
it was sufficient to use our tool
directly on CoNLL files without a GUI.
In the future,
to make the tool accessible to a wider part of the community,
we \revb{plan to integrate }it into an existing
software, \revb{in particular into MMAX2~\cite{Muller2006mmax2} or into }%
BART~(Broscheit, Poesio, Ponzetto, Rodriguez, Romano, Uryupina, Versley, and Zanoli, 2010).
\section{Baseline}
\label{secBaseline}

We have created a baseline for mention detection,
based on the work of Sapena et al.~\shortcite{Sapena2012},
and for coreference resolution,
inspired by Bengtson and Roth~\shortcite{Bengtson2008}.
The baseline \revb{was implemented using }Python and scikit-learn~%
(Pedregosa, Varoquaux, Gramfort, Michel, Thirion, Grisel, Blondel, Prettenhofer,
Weiss, Dubourg, Vanderplas, Passos, Cournapeau, Brucher, Perrot, and Duchesnay, 2011).
\revb{An optional module provides features using FastText word embedding vectors~\cite{Bojanowski2017fasttext}.}
We considered to integrate also the Named Entity Recognition (NER) module
of the ITU-pipeline~\cite{Eryigit2014} because \revb{named entities are }%
not annotated in the Turkish Treebank;
however, we found that the output \revb{of }the web service
changed significantly several times
during the development of the baseline.

To facilitate \revb{replicability }of \revb{our }results,
\revb{the baseline with deactivated FastText module
uses only features that are available in the gold standard of
the METU-Sabanci Turkish Treebank~\cite{Say2004}
and in the Marmara Turkish Coreference Corpus.}

\paragraph{Mention Detection.}
Our Mention Detection baseline {\revamark is rule-based and} marks all
\begin{enumerate}[(i)]
\item noun phrases,
\item pronouns,
\item \revb{named entities, }and
\item capitalized common nouns or proper names that occur two or more times in the document
\end{enumerate}
as mentions.
\revb{As the Turkish Treebank contains no named entity gold annotation,
we heuristically mark all nouns with capitalized lemmas as named entities.
Sentence-initial tokens are always capitalized
and (iv) helps to differentiate sentence-initial named entities
from sentence-initial noun phrase mentions.
The above set of rules (i)--(iv)
reproduces the approach that Sapena et al.~\shortcite{Sapena2012}
described for English mention detection.}

\paragraph{Coreference Resolution.}
Our baseline is similar to the approach described by Bengtson and Roth \shortcite{Bengtson2008}
{\revamark
where coreference chains were predicted with reasonable accuracy
using a small set of features and Support Vector Machines (SVM) \cite{Cortes1995svm}.
}

As input, the baseline uses a set of candidate mentions (either gold or predicted),
lemma information,
and dependency parsing information for obtaining mention heads.
The type of a mention is marked as \emph{pronoun} if the lemma of the
\revb{token is in the list of pronoun lemmas described by
Kılı{\c{c}}aslan et al.~\shortcite{Kilicaslan2009}.}
To separate \emph{proper noun} from \emph{noun phrase} mention types,
we realized our own heuristic which
\bipe{(i)}
\item collects all upper-case tokens not at sentence-initial position,
\item strips case markers, and
\item uses the resulting set of strings to mark all (including sentence-initial) tokens as {proper nouns}.
\eipe
All remaining mentions
\revb{are considered to be }noun phrases.

Based on mention types and head information,
we create the following features
for each \revb{link }$(m_1,m_2)$:
\begin{enumerate}[(i)]
\item \revb{the }type of $m_1$ and type of $m_2$ (2 features),
\item \revb{whether }both mentions are pronouns, proper nouns, or noun phrases (3 features),
\item \revb{whether the }heads of $m_1$ and $m_2$ match, and the same for \revb{the respective }head lemmas (2 features),
\item \revb{whether the }last part of a proper noun is equal in $m_1$ and $m_2$,
\item \revb{whether }$m_1$ is an acronym of $m_2$, and
\item \revb{whether the }head of $m_1$ is a sub-string of \revb{the }head of $m_2$, and the same for \revb{the respective }head lemmas (2 features).
\end{enumerate}
Features (v)\revb{ and (vi) }are asymmetric, that means exchanging $m_1$ and $m_2$ can change the feature value.
For these features we also add the respective reverse direction feature,
as well as the disjunction of features of both directions.
Moreover, we add all possible pairs of features (i)--(ii) and (iii)--(vi)
to allow the machine learning \revb{model }to give separate weight to features (iii)--(vi) per mention type.

{\revbmark
All the above features can be derived from the Turkish Treebank contents
without the need for additional predictions.
To integrate FastText~\cite{Bojanowski2017fasttext} word embedding features
we use ideas from the work of Simova and Uszkoreit (2017).
We use FastText because the vectors are learned from character n-grams of words
which makes them particularly suggestive for usage with the agglutinative morphology of Turkish.
We trained embeddings with default parameters
(100 dimensions, character n-grams of length 3--6, context window of size 5)
and the skip-gram representation on the lowercase dump of Turkish Wikipedia from 1.1.2018.%
\footnote{\protect\revb{\url{https://dumps.wikimedia.org/trwiki/20180101/}}}
The text in that dump contains 73 million tokens and 412459 types.
Word vectors for heads of mentions are looked up directly,
word vectors for mentions are computed as the average over all tokens in the mention.
We add the following embedding features for predicting coreference of a link $(m_1, m_2)$:
\begin{enumerate}[(i)]
\setcounter{enumi}{6}
\item
  cosine similarity between vectors of heads of $m_1$ and $m_2$ (2 features),
\item
  cosine similarity between vectors of mentions $m_1$ and $m_2$ (2 features),
\item
  vectors of heads of $m_1$ and $m_2$ (200 features), and
\item
  vectors of mentions $m_1$ and $m_2$ (200 features).
\end{enumerate}
As done by Simova and Uszkoreit,
we also experimented with Principal Component Analysis (PCA) to transform word vectors into 15 dimensions.
This explains 47\% of the variance in the learned dictionary of word vectors
and creates only 60 features in total for (ix) and (x).}
\smallskip

\subsection{\revb{Coreference Chain Prediction}}
\label{secCorefPrediction}
We implemented two methods for predicting coreference
based on classification (SVC) and regression (SVR).

\emph{SVC}
is based on classification with a linear-kernel SVM \cite{Cortes1995svm}.
Positive examples are mentions and their closest predecessors within all \revb{coreference }chains,
while negative examples are all \revb{non-coreference links }with less than 100 mentions distance.
For predicting \revb{coreference }chains,
we first generate candidate \revb{links }for all mentions
except for \revb{links where the first mention is a pronoun
and the second mention is not a pronoun,
as done by Bengtson and Roth~\shortcite{Bengtson2008}.}
Then, we predict whether \revb{a link is coreferent or not }using the SVM.
Finally, each mention starts in its own \revb{coreference }chain
and we go through mentions from the beginning of the document to the end,
and merge mentions to (potentially several) previous \revb{coreference }chains for all predicted \revb{coreference }links.
We prevent merges that lead to \revb{coreference }chains
with overlapping mentions.

\emph{SVR}
is based on support vector regression with a linear-kernel SVM~%
(Drucker, Burges, Kaufman, Smola, and Vapnik, 1997)
\nocite{Drucker1997svr}%
trained on the same examples as SVC.
For prediction, we generate the same candidate mentions as in SVC.
For building \revb{coreference }chains, we also start with one \revb{coreference }chain per mention,
but this time we use the Best-Link \cite{Bengtson2008} strategy:
we iterate over mentions in order of occurrence in the document,
and merge each mention with at most one predecessor \revb{coreference }chain
if its highest-scored candidate link to a predecessor mention is above 0.1
and if the resulting \revb{coreference }chain does not contain overlapping mentions.
\revb{This threshold was determined in preliminary experiments.
The optimal value can depend on the
difference between the ratio of coreference and non-coreference links
in the training set and in the testing set.)}

In addition to the above,
when predicting coreference on predicted mentions,
we include incorrect mentions predicted on the training documents
to generate negative examples.
We randomly sample as many incorrect mentions as already contained in the gold annotation.
When predicting coreference on gold mentions, we train only on gold mentions.
We balance example weight by class size
(we have significantly more negative examples),
and we use L2 regularization for both SVC and SVR.

\subsection{Evaluation}
We evaluate our baseline using the
CoNLL reference coreference scorer~\cite{Pradhan2014scorer}
and report MUC, B$^3$, CEAF$_m$, CEAF$_e$, BLANC\revb{, and LEA} scores
{\revamark
(see Section~\ref{secCorefMetrics}).
We also use the reference coreference scorer
for obtaining accumulated scores over multiple documents.}

Mention detection is done on the Turkish Treebank data
and does not require learning.
Coreference resolution is done either on gold mentions (GM) or on predicted mentions (PM).
Scores are obtained by leave-one-out cross-validation
on all 33 documents of the corpus, yielding 33 folds.
All scores are given \revb{as }percent\revb{ages}.
For mention detection we report Precision and Recall,
for coreference scores we report only F1.
{\revbmark
We experiment with three feature sets:
\begin{itemize}
\item T includes the features (i)--(vi) which are based on the Turkish Treebank;
\item TF adds to T the features (vii)--(x) with 100-dimensional vectors from FastText; and
\item TP adds to T the features (vii)--(x) with 15-dimensional vectors from applying PCA to FastText embeddings.
\end{itemize}
}

\mybaselinebygenres
Table~\ref{tblBaselineGenre} shows the results of performing mention detection
and coreference with SVC on gold mentions\revb{ (GM) using the T feature set.}
We obtain {\revbmark 88.2\%} Recall %
for mention detection over the whole Treebank.
As expected and as intended, Precision %
is much worse because we expect the coreference resolution step to
eliminate spurious mentions.
Coreference resolution on gold mentions yields a \revb{LEA }score of \revb{57.8\%},
\revb{the more permissive MUC, }B$^3$, CEAF, and BLANC scores are higher.
The worst scores are obtained from genre Other,
which contains a single first-person narrative.
As no other document contains such a narrative,
in cross-validation, the training set contains only documents
that are structurally different from the test set
and a low score is expected.
\revb{We analyse mistakes of the baseline in detail in Section~\ref{secBaselineErrorAnalysis}.}

\mybaselinebymethod
Table~\ref{tblBaselineMethod} shows
overall scores for \revb{predicting coreference on gold mentions using }SVC or SVR
and \revb{for predicting coreference on predicted mentions (PM) using }SVR.
\revb{We show results for all feature sets because for each of the three
sections of the table, the best result is achieved by another feature set.}
SVC/GM is the same setup as in
Table~\ref{tblBaselineGenre}.
\revb{Among the tested configurations,
feature set T yields the best LEA score on gold mentions.
Scores become worse when we add word embedding features;
moreover, high-dimensional embeddings (TF) yield worse scores than
low-dimensional embeddings (TP).}
SVR/GM yields scores \revb{slightly below those of }SVC.
\revb{Differently from SVC, embedding features do not deteriorate the scores.}
The reason for this difference between SVC and SVR is
\revb{the foundational difference for the chain building algorithm of SVC and SVR}:
SVC merges all coreference chains where a coreference link is predicted,
while SVR uses the Best-Link strategy (see Section~\ref{secCorefPrediction}).
\revb{As a consequence, with SVR uses only the highest scoring link
and is not sensitive to variations in scores of links with low confidence,
while SVC considers all links where coreference was predicted with some confidence.
SVC might therefore mistakenly merge many coreference chains.
This increases MUC score
but decreases all more reliable scoring metrics,
as can be observed in Table~\ref{tblBaselineMethod} lines SVC/GM/TF and SVC/GM/TP.}
Naturally, coreference prediction
\revb{yields significantly worse results on PM than on GM,
with a score difference around 30\%.
We analyze possible reasons for this performance difference in the Section~\ref{secBaselineErrorAnalysis}.}

{\revbmark
\paragraph{Further experiments.}
Apart from the results shown in Tables~\ref{tblBaselineGenre} and~\ref{tblBaselineMethod},
we conducted several further experiments.
When using SVC on predicted mentions,
nearly all mentions end up in a single coreference chain
because the presence of many superfluous mentions gives a lot of opportunity
for SVC to mistakenly connect chains.
As discussed above,
SVR does not suffer from this issue due to the Best-Link strategy.
Apart from the features discussed above,
we experimented with using the role of the head
of noun phrase mentions $m_1$ and $m_2$
in the dependency parse graph for predicting coreference of link $(m_1,m_2)$.
Adding this feature causes scores to decrease for all configurations discussed above.
We think this is due to overfitting to the small training corpus
and to the comparatively large number of 58 distinct dependency roles
in the Turkish Treebank.}

\subsection{Error Analysis}
\label{secBaselineErrorAnalysis}
{\revamark
An analysis of missing mentions in Mention Detection
uncovered several examples of problems with tokens that contain a derivation boundary.
Such tokens are, for example, adjectivised verbs,
which are not counted as being part of a noun phrase although they sometimes should be.
A concrete example is \quo{sona eren bin yıl}
(\quo{thousand years that are coming to an end})
where \quo{sona eren} (\quo{coming to an end}) is a single token in the Treebank
that is a modifier of \quo{bin yıl} (\quo{thousand years}).
This token is an adjectivised verb and contains a derivation boundary.
A possibility for improving mention detection
could be to split tokens with derivation boundaries into multiple tokens:
a base lexical token and potentially multiple virtual functional tokens
for deciding mention boundaries.
However, this effort would exceed the construction of a baseline
and we consider such improvements as future work.

{\revbmark
A quantitative analysis of mistakes with gold mentions showed
that 58\% of predicted coreference chains are missing at least one mention,
the remaining 42\% of wrongly predicted coreference chains
contain mentions from at least two gold standard coreference chains.
These mistakes show a balance between predicting too many links and too few links.
Improving predictions on gold mentions, therefore, likely requires new features
or more expressive (i.e., non-linear) combinations of existing features
in the machine learning model
which could be achieved by feature engineering or by applying deep learning methods.}
An \revb{inspection of }mistakes showed
that the genres Travel and Other obtain a worse score because there is a frequent
alternation between first and third person,
without an implicit change of the speaker,
such that often \quo{biz} (\quo{we}) and \quo{siz} (\quo{you}) corefer
although the person changes.
A frequent source of mistakes in the Novel genre is the usage of specific
family titles like \quo{hala} (\quo{aunt on the father's side})
and \quo{abi} (\quo{older brother})
which allow \revb{inferences about }specific coreferen\revb{tial mentions to be made }through reasoning about previously explained family situations.
However, our baseline system is not aware of the meaning of these words
and we consider the introduction of such reasoning into
Turkish coreference resolution as future work.}

{\revbmark
A quantitative analysis of mistakes with predicted mentions showed
that 58\% of predicted coreference chains contain only spurious mentions
which are not part of any coreference chain in the gold standard.
Moreover, 17\% of predicted coreference chains are missing some mention,
14\% of predicted coreference chains contain mentions that are not part of the gold standard,
and only 12\% of predicted coreference chains merge two or more gold standard chains.
For improving coreference prediction on predicted mentions,
we think that developing a predictor for distinguishing
between specific and generic mentions of a common noun will be important
to avoid superfluous mentions and therefore superfluous coreference links.
We think this is a promising future direction because superfluous mentions
account for 72\% (58\%+14\%) of mistakes in chains.}

As this is only a baseline,
we did not include more sophisticated features
described by Bengtson and Roth~\shortcite{Bengtson2008}.
For example, semantic features based on WordNet~%
(Miller, 1995;
Bilgin, {\c{C}}etinoğlu, and Oflazer, 2004)
could rule out certain predicted mentions as relevant
and thus could improve Precision of the baseline.

\section{Conclusion}
\label{secConclusion}
{\revbmark
We have presented the Marmara Turkish Coreference Corpus,
the first Turkish coreference corpus,
based on the de facto standard Turkish Treebank.
We also presented a baseline system for mention detection and coreference prediction.

Our corpus has coreference annotated on the token level.
This has several reasons:
the relevant possessive and personal pronoun suffixes
are not accessible as markables in the Turkish Treebank,
a morpheme-based coreference annotation would increase the complexity
of the annotation task,
and it would require annotators
with a higher level of expertise than available in our study.}
For a future annotation project it could be interesting to extend annotations to include coreference links\revb{ to morphemes}.
\revb{This would require to extend the Turkish Treebank so that inflectional groups
are structured further into morphemes to make them accessible as markables
(see Section~\ref{secTurkishCoref}).}
\revb{For }scoring with the reference scorer tool \revb{it would be necessary to }%
develop a CoNLL representation for Turkish where tokens are split within IGs.

{\revbmark
The (relative) simplicity of the annotation task
and the high number of annotators that were involved in this project
(i.e., at least eight annotators for each document)
allowed us to perform adjudication automatically.
Our analysis of the adjudication process shows,
that 74\% of all annotated coreference chains fully correspond to a
coreference chain in the gold standard,
and the majority of non-perfectly annotated chains either misses a few mentions
or contains a few spurious mentions (see Section~\ref{secAnnotatorMistakes}).
Therefore, we are confident that the annotation and adjudication process
has yielded a reliable coreference gold standard.}

{\revbmark
The baseline demonstrates,
that the created gold standard is consistent in itself
and permits prediction of mentions and coreference links
using existing methods from literature.
}
To improve the mention detection baseline,
information about appositives as well as finding a way
to filter out generic mentions would be useful.
To improve the coreference resolution baseline,
adding more complex features by
integrating Turkish WordNet~\cite{Bilgin2004wordnet},
Turkish NER~\cite{Seker2012},
and Turkish WSD~%
(İlgen, Adalı, and Tantuğ, 2012)
could be helpful.
For a full processing pipeline from plain text to coreference
annotations,
\revb{the baseline described here derives features from}\
morphological analysis and disambiguation~%
(Sak, G{\"{u}}ng{\"{o}}r, and Sara{\c{c}}lar, 2007),
and \revb{from }dependency parsing~%
(Eryiğit, Nivre, and Oflazer, 2008).
Available tools \revb{for these tasks }are the ITU-pipeline~\cite{Eryigit2014}
and the older Zemberek system~\cite{Akin2007zemberek}.
\revb{For providing a replicable evaluation of the baseline,
we use only features that are present in the Turkish Treebank
(and optionally word embedding vectors).}

{\revbmark
Orthogonal to our baseline,
it would be interesting to build a joint mention and coreference prediction system
based on deep learning with only word embeddings as the main input,
similar to the systems of Wu et al. \shortcite{Wu2017deepcoref}
and Lee et al. \shortcite{Lee2017neuralcoref}.}

\section*{Acknowledgements}
We are grateful to Kemal Oflazer and Bilge Say
for support about the METU-Sabanci Turkish Treebank,
and to Dilek Küçük and Savaş Yıldırım for support about their papers and datasets.
\revb{We are grateful to the reviewers for their constructive comments.}

This work has been supported by
The Scientific and Technological Research Council of Turkey
(TUBITAK) under grant agreements 114E430 and 114E777.

Author contributions are as follows (in chronological order):
creating documents from Treebank fragments~(B.G.S),
evaluating GATE and BRAT annotation software~(B.G.S., A.P.),
creating the annotation manual~(P.S., K.C., F.T., B.G.S.),
performing manual adjudication in \phaseone~(K.C., B.G.S., A.P., H.E.K.),
annotating mentions for those documents in \phasetwo\ that were
not annotated in \phaseone~(K.C., H.E.K.),
revising the manuscript as an English native-speaker~(A.H.K.),
writing the baseline software~(P.S., F.T., A.P.),
managing the project,
performing baseline experiments,
analysing data and results,
writing and revising the manuscript~(P.S).

\bibliographystyle{apalike}

\ifinlineref

\input{references.sty}

\else
\bibliography{library-backup} %
\fi

\end{document}